%% file: aaai2026.tex
\documentclass[letterpaper]{article} 
\usepackage{aaai2026}  
\usepackage{times}  
\usepackage{helvet}  
\usepackage{courier}  
\usepackage[hyphens]{url}  
\usepackage{graphicx} 
\usepackage{natbib}  
\usepackage{caption} 
\usepackage{multirow}
\usepackage{booktabs}
\usepackage{pifont}
\usepackage{caption}
\usepackage{array}
\usepackage{colortbl}
\usepackage{xcolor}         
\usepackage{subcaption}
\usepackage{amssymb}

\definecolor{baseline}{rgb}{0.0, 0.0, 0.5}
\definecolor{darkgreen}{RGB}{0, 100, 0} 
\newcommand{\cmark}{\textcolor{darkgreen}{\ding{51}}} 
\newcommand{\xmark}{\textcolor{red}{\ding{55}}}   
\usepackage{algorithm}
\usepackage{amsmath}
\usepackage{algpseudocode}
\algrenewcommand\algorithmicrequire{\textbf{Input:}}
\algrenewcommand\algorithmicensure{\textbf{Output:}}
\makeatletter
\renewcommand{\alglinenumber}[1]{}   
\makeatother
\usepackage{newfloat}
\usepackage{listings}
\DeclareCaptionStyle{ruled}{labelfont=normalfont,labelsep=colon,strut=off} 
\floatstyle{ruled}
\newfloat{listing}{tb}{lst}{}
\floatname{listing}{Listing}
\title{Meta-Learned Adaptive Optimization for Robust Human Mesh Recovery with Uncertainty-Aware Parameter Updates}
\author{
    Shaurjya Mandal,
    Nutan Sharma,
    John Galeotti
}

\begin{document}

\maketitle


\begin{abstract}
    Human mesh recovery from single images remains challenging due to inherent depth ambiguity and limited generalization across domains. While recent methods combine regression and optimization approaches, they struggle with poor initialization for test-time refinement and inefficient parameter updates during optimization. We propose a novel meta-learning framework that trains models to produce optimization-friendly initializations while incorporating uncertainty-aware adaptive updates during test-time refinement. Our approach introduces three key innovations: (1) a meta-learning strategy that simulates test-time optimization during training to learn better parameter initializations, (2) a selective parameter caching mechanism that identifies and freezes converged joints to reduce computational overhead, and (3) distribution-based adaptive updates that sample parameter changes from learned distributions, enabling robust exploration while quantifying uncertainty. Additionally, we employ stochastic approximation techniques to handle intractable gradients in complex loss landscapes. Extensive experiments on standard benchmarks demonstrate that our method achieves state-of-the-art performance, reducing MPJPE by 10.3 on 3DPW and 8.0 on Human3.6M compared to strong baselines. Our approach shows superior domain adaptation capabilities with minimal performance degradation across different environmental conditions, while providing meaningful uncertainty estimates that correlate with actual prediction errors. Combining meta-learning and adaptive optimization enables accurate mesh recovery and robust generalization to challenging scenarios.
\end{abstract}

\section{Introduction} \label{sec:introduction}

\input{chapters/introduction}

\section{Related Works} \label{sec:related}
\input{chapters/related_works}

\section{Methodology} \label{sec:method}
\input{chapters/methodology}

\section{Experiments} \label{sec:experiments}
\input{chapters/experiments}

\section{Conclusion} \label{sec:conclusion}
\input{chapters/conclusion}

\bibliography{aaai2026.bib}

\clearpage


\input{chapters/checklist}

\end{document}

%% file: chapters/introduction.tex
Reconstructing a complete 3D human body mesh from a single RGB image remains one of the most challenging problems in computer vision, with applications spanning augmented reality, human-computer interaction, and motion analysis. The task, commonly referred to as Human Mesh Recovery (HMR), is inherently ill-posed due to depth ambiguity---multiple distinct 3D poses can project to identical 2D observations, creating fundamental uncertainty in the reconstruction process. Early works addressed this ambiguity through parametric human body models, most notably the Skinned Multi-Person Linear (SMPL) model~\cite{loper2015smpl}, which provides a compact representation of human geometry through pose and shape parameters. This parameterization effectively constrains the solution space to anatomically plausible configurations, enabling tractable optimization even from ambiguous 2D evidence. Contemporary HMR methods broadly fall into two paradigms: \textbf{optimization-based} and \textbf{regression-based} approaches. Optimization-based methods iteratively fit parametric models to image observations (2D joint detections), often achieving precise alignment between projected models and visual evidence. However, these approaches suffer from computational inefficiency and sensitivity to initialization, as high-dimensional parameter optimization remains challenging. In contrast, regression-based methods employ deep neural networks to directly predict SMPL parameters from input pixels in a single forward pass. While computationally efficient, these approaches rely heavily on large-scale training data and struggle with out-of-distribution inputs, failing to explicitly handle the inherent ambiguity in novel scenarios.

Recent research has explored hybrid methodologies that combine the strengths of both paradigms. Notable examples include SPIN \cite{kolotouros2019learning} (SMPL oPtimization IN the loop), which integrates optimization steps within the training loop to create self-improving regressors. These approaches demonstrate that incorporating optimization paradigms can significantly enhance pose accuracy while maintaining computational efficiency.

Another research direction addresses ambiguity through \textbf{multi-hypothesis prediction}. Methods by Biggs \cite{biggs2020multiple}, Li and Lee \cite{li2021mixture}, and Kolotouros \cite{kolotouros2021probabilistic} (ProHMR) generate multiple plausible pose hypotheses or full probability distributions over possible configurations. While these probabilistic approaches better capture inherent uncertainty, they complicate downstream processing and pose challenges in selecting optimal predictions from multiple candidates.

Simultaneously, \textbf{meta-learning techniques} have emerged as powerful tools for improving generalization. Recent work by Nie \cite{nie2022metahmr} (MetaHMR) incorporates test-time optimization into training via meta-learning frameworks, producing models that adapt more effectively to novel conditions during inference.

\subsection{Key Limitations of Existing Methods}

Despite these advances, current approaches face several critical limitations:

\begin{enumerate}
    \item \textbf{Uncertainty-Accuracy Trade-off}: Multi-hypothesis methods provide uncertainty quantification but complicate inference, while single-output methods lack principled uncertainty handling.
    
    \item \textbf{Computational Inefficiency}: Optimization-based refinement often involves redundant parameter updates, particularly for already-converged joints.
    
    \item \textbf{Limited Adaptability}: Pure regression models struggle with domain shifts, while optimization methods remain sensitive to initialization quality.
    
    \item \textbf{Gradient Intractability}: Complex loss functions and discrete operations can render analytical gradients unreliable or intractable.
\end{enumerate}

\begin{figure*}[t]
\centering
\includegraphics[width=\textwidth]{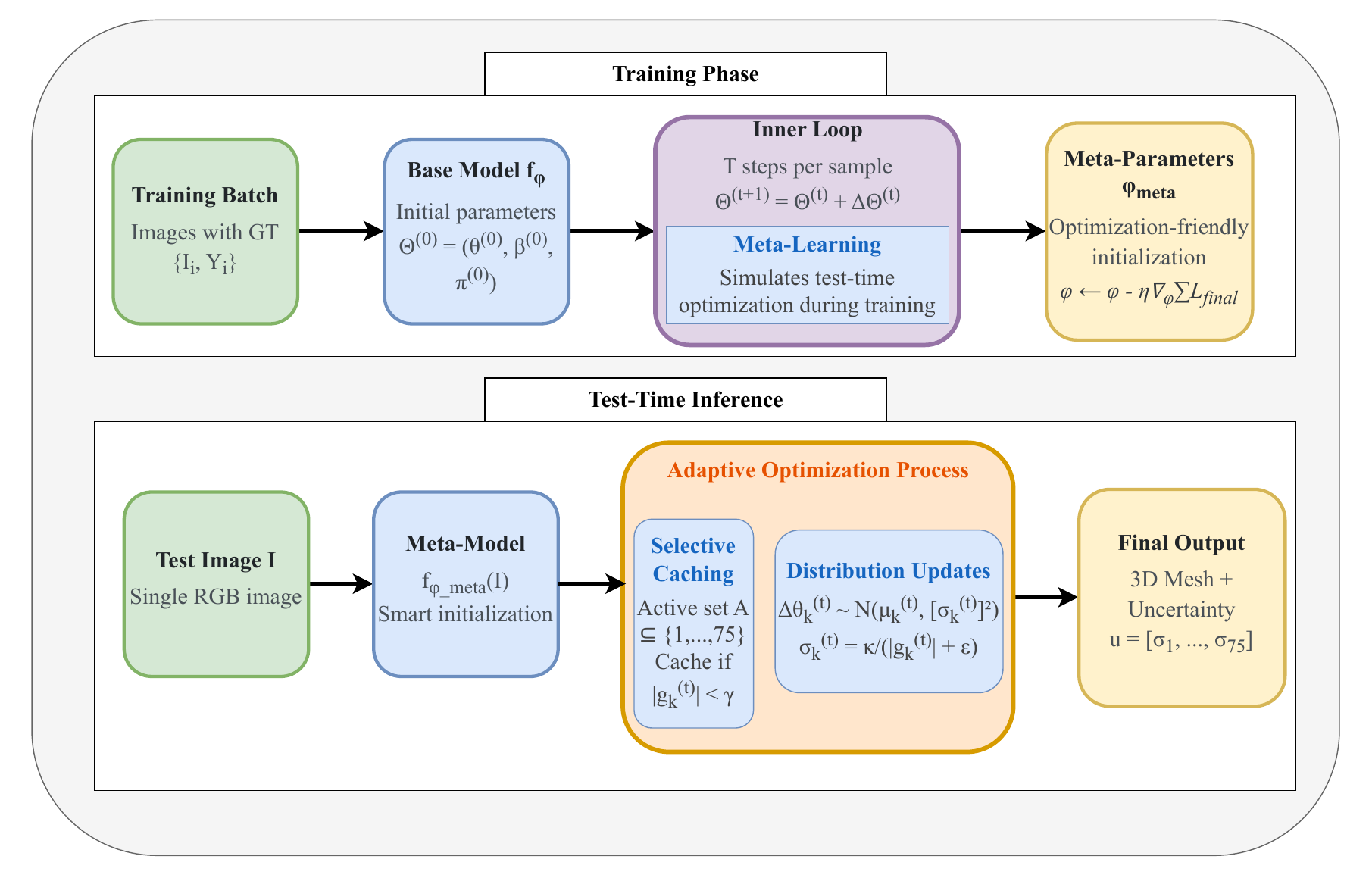}
\caption{\textbf{Overview of our meta-learned adaptive optimization framework for human mesh recovery.}}
\label{fig:main}
\end{figure*}

\subsection{Our Contribution}

We propose a novel HMR framework (Figure~\ref{fig:main}) that addresses these limitations through three key innovations while maintaining deterministic single-output simplicity:

\begin{enumerate}
 \item\textbf{Meta-Learned Initialization}: We employ meta-learning to train optimization-friendly initial parameter predictions that enable rapid convergence across diverse data distributions, even for unseen scenarios.

 \item\textbf{Selective Parameter Caching}: We introduce an intelligent caching mechanism that identifies and temporarily freezes joints with negligible expected updates, eliminating redundant computations and preventing optimization instabilities.

 \item\textbf{Distribution-Based Adaptive Updates}: We formulate pose refinement as sampling from learned distributions over joint angle adjustments, where distribution parameters adapt based on local optimization states. This provides implicit uncertainty quantification while enabling robust handling of non-smooth loss landscapes through stochastic approximation techniques.
\end{enumerate}

Our approach delivers accurate pose estimates with meaningful uncertainty measures, demonstrating superior resilience to domain shifts compared to existing methods. We validate our contributions on standard benchmarks and introduce cross-dataset protocols to evaluate out-of-domain performance, showing consistent improvements in both accuracy and adaptability.

%% file: chapters/related_works.tex
We review existing approaches to human mesh recovery, focusing on the evolution from pure regression to hybrid optimization methods, and the emerging role of meta-learning and uncertainty modeling.

\subsection{Regression-Based Methods}

Early neural approaches to HMR directly regress parametric body model parameters from RGB images. The seminal work by Kanazawa \cite{kanazawa2018end} introduces a CNN-based framework that predicts SMPL pose and shape parameters in a single forward pass. This paradigm enables fast inference and has been widely adopted with various architectural improvements \cite{sun2019human,kocabas2020vibe}. While computationally efficient, pure regression methods suffer from inherent limitations. They tend to produce averaged solutions when multiple plausible poses exist, struggling with ambiguous scenarios such as occlusions or rare viewpoints. Additionally, these methods require extensive labeled data and exhibit poor generalization to out-of-distribution inputs \cite{zhang2021pymaf}.

\subsection{Optimization-Based Methods}

Optimization approaches iteratively fit parametric models to image observations by minimizing carefully designed objective functions. SMPLify \cite{bogo2016keep} exemplifies this paradigm, optimizing SMPL parameters to match detected 2D keypoints through iterative refinement. These methods achieve precise image-model alignment and naturally handle weak supervision scenarios. However, optimization-based approaches face significant computational overhead, requiring dozens of iterations per image. They are also sensitive to initialization quality and prone to local minima, necessitating careful objective design and hyperparameter tuning \cite{pavlakos2019expressive}.

\subsection{Hybrid Approaches}

Recent works combine regression and optimization to leverage the strengths of both paradigms. SPIN \cite{kolotouros2019learning} tightly integrates optimization into the training loop: neural network predictions initialize in-training optimization steps, with the refined results supervising the network. This creates a self-improving system that significantly outperforms purely regression-based methods. Extensions like EFT \cite{joo2020exemplar} and PyMAF \cite{zhang2021pymaf} further demonstrate that incorporating optimization feedback during training substantially improves accuracy. Song \cite{song2020human} propose learning the gradient descent process itself through differentiable reprojection errors.

\subsection{Meta-Learning for Adaptation}

Meta-learning techniques have recently been applied to improve HMR generalization. MetaHMR \cite{nie2022metahmr} incorporates test-time optimization into training iterations, teaching models to adapt quickly to new scenarios. Their approach addresses objective misalignment between training and test-time optimization through dual-network architectures. Inspired by Model-Agnostic Meta-Learning (MAML) \cite{finn2017model}, these methods optimize for initializations that enable rapid adaptation with minimal gradient steps. Our work extends this philosophy by combining meta-learning with stochastic optimization strategies.

\subsection{Uncertainty and Multi-Hypothesis Methods}

Given HMR's ill-posed nature, several approaches explicitly model prediction uncertainty. ProHMR \cite{kolotouros2021probabilistic} uses conditional normalizing flows to represent complex pose distributions, enabling multiple plausible reconstructions per image. Biggs \cite{biggs2020multiple} extend standard architectures to output multiple discrete hypotheses, capturing diverse pose interpretations. While these methods provide explicit uncertainty quantification, they complicate downstream processing and require careful hypothesis selection. In contrast, our approach maintains deterministic single outputs while leveraging uncertainty internally during optimization.

Our method uniquely integrates meta-learning, stochastic optimization, and selective parameter updates. Unlike existing approaches that treat optimization steps as deterministic processes, we formulate updates as samples from learned distributions that adapt based on local optimization states. The selective caching mechanism further distinguishes our work by dynamically freezing converged parameters, reducing computational overhead while preventing optimization instabilities. To our knowledge, no existing HMR method combines distribution-based adaptive updates with parameter caching in a meta-learned framework, representing a novel integration of established concepts for improved performance and efficiency.

%% file: chapters/methodology.tex
\subsection{Problem Formulation}

We estimate 3D human body mesh from a single RGB image $I$ by predicting parameters of the SMPL model \cite{loper2015smpl}. SMPL represents human body geometry through pose parameters $\boldsymbol{\theta} \in \mathbb{R}^{72}$ (24 joint rotations in axis-angle format) and shape parameters $\boldsymbol{\beta} \in \mathbb{R}^{10}$. The model defines a differentiable function $M(\boldsymbol{\theta}, \boldsymbol{\beta}) \rightarrow \mathbb{R}^{6890 \times 3}$ that outputs mesh vertices, with 3D joint locations obtained via linear regressor $J = W \cdot M(\boldsymbol{\theta}, \boldsymbol{\beta})$ where $W \in \mathbb{R}^{24 \times 6890}$.

We also estimate weak-perspective camera parameters $\boldsymbol{\pi} = [s, t_x, t_y]^T$ where $s$ denotes scale and $(t_x, t_y)$ translation. Given image $I$, our goal is to predict $\hat{\boldsymbol{\theta}}, \hat{\boldsymbol{\beta}}, \hat{\boldsymbol{\pi}}$ such that the projected mesh aligns with image evidence.

Our approach consists of two stages: (1) initial prediction via meta-learned base model $f_{\phi}(I)$ with parameters $\phi$, producing $(\boldsymbol{\theta}^{(0)}, \boldsymbol{\beta}^{(0)}, \boldsymbol{\pi}^{(0)})$, and (2) iterative refinement minimizing energy function (Eq.~\ref{eq:energy}):

\begin{equation}
\label{eq:energy}
\begin{aligned}
E(\boldsymbol{\theta}, \boldsymbol{\beta}, \boldsymbol{\pi}; I) &= \sum_{j=1}^{J} w_j \left\|\Pi_{\boldsymbol{\pi}}\big([W \cdot M(\boldsymbol{\theta}, \boldsymbol{\beta})]_j\big) - u_j\right\|^2 \\
&\quad + \lambda_{\text{pose}} L_{\text{pose}}(\boldsymbol{\theta}) + \lambda_{\text{shape}} L_{\text{shape}}(\boldsymbol{\beta})
\end{aligned}
\end{equation}

where $u_j \in \mathbb{R}^2$ is the $j$-th detected 2D keypoint, $w_j$ its confidence weight, $\Pi_{\boldsymbol{\pi}}: \mathbb{R}^3 \rightarrow \mathbb{R}^2$ the projection function parameterized by $\boldsymbol{\pi}$, and $L_{\text{pose}}, L_{\text{shape}}$ are regularization terms.

\subsection{Meta-Learning Framework}

We formulate HMR training as a meta-learning problem where each image represents a task requiring rapid adaptation. The framework employs inner-loop refinement and outer-loop parameter updates following MAML principles~\cite{finn2017model}.

For training batch $\{(I_i, Y_i)\}_{i=1}^N$ where $Y_i$ denotes ground truth annotations, the base model produces initial estimates $\Theta_i^{(0)} = (\boldsymbol{\theta}_i^{(0)}, \boldsymbol{\beta}_i^{(0)}, \boldsymbol{\pi}_i^{(0)}) = f_{\phi}(I_i)$. We then simulate test-time refinement through $T$ optimization steps using the update rule (Eq.~\ref{eq:inner_update}):

\begin{equation}
\Theta_i^{(t+1)} = \Theta_i^{(t)} + \Delta \Theta_i^{(t)}
\label{eq:inner_update}
\end{equation}

where $\Delta \Theta_i^{(t)}$ represents parameter updates computed via our distribution-based strategy (\S\ref{sec:adaptive_updates}). Note that $\Theta_i^{(t)} = (\boldsymbol{\theta}_i^{(t)}, \boldsymbol{\beta}_i^{(t)}, \boldsymbol{\pi}_i^{(t)})$ includes all parameters for image $i$ at iteration $t$.

The outer-loop update optimizes base model parameters via gradient descent (Eq.~\ref{eq:outer_update}):

\begin{equation}
\phi \leftarrow \phi - \eta \nabla_{\phi} \sum_{i=1}^N L_{\text{final}}(\Theta_i^{(T)}, Y_i)
\label{eq:outer_update}
\end{equation}

where $\eta$ denotes outer-loop learning rate and $L_{\text{final}}$ measures final prediction error (e.g., 3D vertex loss).

We employ dual supervision combining final and intermediate losses (Eq.~\ref{eq:dual_loss}):

\begin{equation}
\label{eq:dual_loss}
\begin{aligned}
L_{\text{train}} &= \sum_{i=1}^{N} \Big( L_{\text{final}}(\Theta_i^{(T)}, Y_i)
\\ &\quad + \sum_{t=0}^{T-1} \alpha_t\, L_{\text{proj}}(\Theta_i^{(t)}; I_i) \Big)
\end{aligned}
\end{equation}

where $L_{\text{proj}}$ represents 2D reprojection loss and $\alpha_t$ are decay weights ensuring inner-loop gradients remain meaningful.

\subsection{Selective Parameter Caching}

High-dimensional optimization often involves redundant parameter updates. We introduce selective caching that maintains an active parameter set $\mathcal{A} \subseteq \{1, 2, \ldots, 75\}$ for all parameters requiring updates (72 pose + 3 camera parameters).

Initially, $\mathcal{A} = \{1, 2, \ldots, 75\}$. At iteration $t$, we compute gradients $g_k^{(t)} = \frac{\partial E}{\partial \theta_k}\big|_{\Theta^{(t)}}$ for parameters $k \in \{1,\ldots,75\}$ and define update decisions using Eq.~\ref{eq:update_decision}:

\begin{equation}
u_j^{(t)} = \begin{cases}
1, & \text{if } j \in \mathcal{A} \text{ and } |g_j^{(t)}| > \gamma \\
0, & \text{otherwise}
\end{cases}
\label{eq:update_decision}
\end{equation}

where $\gamma$ represents the significance threshold. Parameters with $u_k^{(t)} = 0$ are cached (removed from $\mathcal{A}$), yielding selective updates (Eq.~\ref{eq:selective_update}):

\begin{equation}
\theta_k^{(t+1)} = \begin{cases}
\theta_k^{(t)} + \Delta\theta_k^{(t)}, & \text{if } u_k^{(t)} = 1 \\
\theta_k^{(t)}, & \text{if } u_k^{(t)} = 0
\end{cases}
\label{eq:selective_update}
\end{equation}

This mechanism provides computational efficiency by reducing optimization dimensionality and stability by preventing oscillations in converged parameters.

\begin{algorithm}[t]
\small
\caption{Meta-Training Procedure}
\label{alg:training}
\begin{algorithmic}
\Require Training data $\{(I_i,Y_i)\}_{i=1}^N$, model $f_{\phi}$, steps $T$
\While{not converged}
  \State Sample batch $\{(I_i,Y_i)\}_{i=1}^N$
  \ForAll{$(I_i,Y_i)$}
    \State $\Theta_i^{(0)} \gets f_{\phi}(I_i)$
    \State $\mathcal{A} \gets \{1,\ldots,75\}$; $P_k^{(0)} \gets \mathcal{N}(0,\sigma_{\max}^2)$
    \For{$t = 0$ to $T-1$}
      \ForAll{$k \in \mathcal{A}$}
        \State $g_k^{(t)} \gets \frac{\partial E}{\partial \theta_k}$
        \If{$|g_k^{(t)}| < \gamma$}
          \State $\mathcal{A} \gets \mathcal{A} \setminus \{k\}$ \textbf{continue}
        \EndIf
        \State Update $(\mu_k^{(t)}, \sigma_k^{(t)})$ per Eqs.~\eqref{eq:step_size}--\eqref{eq:momentum_sigma}
        \State Sample $\Delta\theta_k^{(t)} \sim \mathcal{N}(\mu_k^{(t)}, [\sigma_k^{(t)}]^2)$
      \EndFor
      \State $\Theta_i^{(t+1)} \gets \Theta_i^{(t)} + \Delta\Theta_i^{(t)}$
    \EndFor
  \EndFor
  \State $\phi \gets \phi - \eta \nabla_{\phi}\frac{1}{N}\sum_{i=1}^N L_{\text{final}}(\Theta_i^{(T)}, Y_i)$
\EndWhile
\end{algorithmic}
\end{algorithm}

\begin{algorithm}[t]
\small
\caption{Test-Time Inference}
\label{alg:inference}
\begin{algorithmic}
\Require Trained model $f_{\phi^{*}}$, image $I$, max steps $T_{\max}$
\State $\Theta^{(0)} \gets f_{\phi^*}(I)$
\State $\mathcal{A} \gets \{1,\ldots,75\}$; $P_k^{(0)} \gets \mathcal{N}(0,\sigma_{\max}^2)$
\For{$t = 0$ to $T_{\max}-1$}
  \ForAll{$k \in \mathcal{A}$}
    \State Compute $g_k^{(t)}$ (analytical or SPSA)
    \If{$|g_k^{(t)}| < \gamma$}
      \State $\mathcal{A} \gets \mathcal{A} \setminus \{k\}$ \textbf{continue}
    \EndIf
    \State Update $(\mu_k^{(t)}, \sigma_k^{(t)})$
    \State Sample $\Delta\theta_k^{(t)} \sim \mathcal{N}(\mu_k^{(t)}, [\sigma_k^{(t)}]^2)$
  \EndFor
  \State $\Theta^{(t+1)} \gets \Theta^{(t)} + \Delta\Theta^{(t)}$
  \If{$\mathcal{A} = \varnothing$ \textbf{ or } $\|\Delta\Theta^{(t)}\| < \epsilon$}
    \State \textbf{break}
  \EndIf
\EndFor
\State \Return $\Theta^{(t+1)}$, uncertainty $\{\sigma_k^{(t)}\}$
\end{algorithmic}
\end{algorithm}

\subsection{Distribution-Based Adaptive Updates}
\label{sec:adaptive_updates}

Our key innovation treats parameter updates as samples from learned distributions rather than deterministic gradient steps. For tractability, we factorize the joint update distribution as shown in Eq.~\ref{eq:factorized_dist}:

\begin{equation}
P^{(t)}(\Delta \Theta) = \prod_{j=1}^{72} P_j^{(t)}(\Delta \theta_j)
\label{eq:factorized_dist}
\end{equation}

Each $P_k^{(t)}$ follows a Gaussian distribution $\mathcal{N}(\mu_k^{(t)}, [\sigma_k^{(t)}]^2)$ where $\mu_k^{(t)}$ represents the expected update direction and $\sigma_k^{(t)}$ captures uncertainty.

\textbf{Mean Update}: We set $\mu_k^{(t)} = -\alpha_k^{(t)} g_k^{(t)}$ where $\alpha_k^{(t)}$ is an adaptive step size. We implement momentum-based adaptation (Eq.~\ref{eq:step_size}):

\begin{equation}
\alpha_k^{(t)} = \alpha_{\text{base}} \cdot \exp\left(\frac{\text{success\_rate}_k^{(t)} - 0.5}{0.1}\right)
\label{eq:step_size}
\end{equation}

where $\text{success\_rate}_k^{(t)}$ tracks recent loss reduction for parameter $k$.

\textbf{Variance Adaptation}: We set variance inversely proportional to gradient confidence (Eq.~\ref{eq:variance}):

\begin{equation}
\sigma_k^{(t)} = \max\left(\sigma_{\min}, \min\left(\sigma_{\max}, \frac{\kappa}{|g_k^{(t)}| + \varepsilon}\right)\right)
\label{eq:variance}
\end{equation}

where $\kappa, \varepsilon, \sigma_{\min}, \sigma_{\max}$ are hyperparameters. Large gradients yield small variance (trust the direction), while small gradients encourage exploration.

Distribution parameters evolve with momentum (Eqs.~\ref{eq:momentum_mu}-\ref{eq:momentum_sigma}):

\begin{align}
\mu_k^{(t+1)} &= \beta_{\mu} \mu_k^{(t)} + (1-\beta_{\mu})(-\alpha_k^{(t)} g_k^{(t)}) \label{eq:momentum_mu} \\
\sigma_k^{(t+1)} &= \beta_{\sigma} \sigma_k^{(t)} + (1-\beta_{\sigma}) \tilde{\sigma}_k^{(t)} \label{eq:momentum_sigma}
\end{align}

where $\tilde{\sigma}_k^{(t)}$ follows Eq.~\ref{eq:variance} and $\beta_{\mu}, \beta_{\sigma} \in [0,1)$ are momentum coefficients.

\textbf{Stochastic Approximation}: When analytical gradients are intractable, we employ Simultaneous Perturbation Stochastic Approximation (SPSA) as in Eq.~\ref{eq:spsa}:

\begin{equation}
\hat{g}_j^{(t)} = \frac{E(\ldots, \theta_j + \delta, \ldots) - E(\ldots, \theta_j - \delta, \ldots)}{2\delta}
\label{eq:spsa}
\end{equation}

where $\delta$ is a small random perturbation.

\subsection{Uncertainty Quantification}

Our method provides uncertainty estimates through final distribution parameters. High variance $\sigma_k^{(\text{final})}$ indicates ambiguous parameter configurations, while low variance suggests confident estimates. We output uncertainty vector $\mathbf{u} = [\sigma_1^{(\text{final})}, \ldots, \sigma_{75}^{(\text{final})}]^T$ alongside parameter predictions.

These estimates enable confidence-based downstream processing and correlate well with actual prediction errors in our experiments.

\subsection{Training and Inference Procedures}

Algorithm~\ref{alg:training} summarizes meta-training, while Algorithm~\ref{alg:inference} details test-time inference. We typically use $T=3\text{-}5$ inner-loop steps during training and allow up to 20 iterations at test time, though convergence usually occurs within 10 steps.

%% file: chapters/experiments.tex
\subsection{Datasets and Evaluation Metrics}

Following established protocols~\cite{lin2021end,zhang2021pymaf}, we train our method on a mixture of four datasets: Human3.6M~\cite{ionescu2013human3}, MPI-INF-3DHP~\cite{mehta2017monocular}, COCO~\cite{lin2014microsoft}, and MPII~\cite{andriluka20142d}. We evaluate on 3DPW~\cite{vonmarcard2018recovering} and Human3.6M test sets. We employ standard evaluation metrics: MPJPE (Mean Per Joint Position Error), PA-MPJPE (Procrustes-aligned MPJPE), and PVE (Per-Vertex Error), all measured in millimeters. Lower values indicate better performance.

\begin{table*}[t]
\centering
\caption{Quantitative comparison with state-of-the-art methods on 3DPW~\cite{vonmarcard2018recovering} and Human3.6M~\cite{ionescu2013human3} datasets. Best results in \textbf{bold}.}
\label{tab:main_results}
\resizebox{\textwidth}{!}{
\begin{tabular}{l|l|l|ccc|cc}
\hline
\multirow{2}{*}{\textbf{Method}} & \multirow{2}{*}{\textbf{Type}} & \multirow{2}{*}{\textbf{Backbone}} & \multicolumn{3}{c|}{\textbf{3DPW}} & \multicolumn{2}{c}{\textbf{Human3.6M}} \\
 & & & \textbf{MPJPE↓} & \textbf{PA-MPJPE↓} & \textbf{PVE↓} & \textbf{MPJPE↓} & \textbf{PA-MPJPE↓} \\
\hline
\multicolumn{8}{c}{\textit{Regression-based}} \\
\hline
HMR~\cite{kanazawa2018end} & Reg & ResNet-50 & 130.0 & 81.3 & - & 88.0 & 56.8 \\
CLIFF~\cite{li2022cliff} & Reg & HRNet-W48 & 69.0 & 43.0 & 81.2 & 47.1 & 32.7 \\
PyMAF~\cite{zhang2021pymaf} & Reg & HRNet-W48 & 74.2 & 45.3 & 87.0 & 54.2 & 37.2 \\
METRO~\cite{lin2021end} & Reg & HRNet-W64 & 77.1 & 47.9 & 88.2 & 54.0 & 36.7 \\
FastMETRO~\cite{cho2022cross} & Reg & HRNet-W64 & 73.5 & 44.6 & 84.1 & 52.2 & 33.7 \\
ProPose~\cite{fang2023learning} & Reg & HRNet-W48 & 68.3 & 40.6 & 79.4 & 45.7 & 29.1 \\
\hline
\multicolumn{8}{c}{\textit{Optimization-based}} \\
\hline
SMPLify~\cite{bogo2016keep} & Opt & - & 96.9 & 59.2 & 116.4 & 62.5 & 41.1 \\
SPIN~\cite{kolotouros2019learning} & Hybrid & ResNet-50 & 92.1 & 54.2 & 109.3 & 63.0 & 40.9 \\
EFT~\cite{joo2021exemplar} & Test-Opt & ResNet-50 & 85.1 & 52.2 & 98.7 & 63.2 & 43.8 \\
HybrIK~\cite{li2021hybrik} & Hybrid & ResNet-34 & 74.1 & 45.0 & 86.5 & 55.4 & 33.6 \\
ReFit~\cite{wang2023refit} & Test-Opt & HRNet-W48 & 65.8 & 41.0 & - & 48.4 & 32.2 \\
PLIKS~\cite{shetty2023pliks} & Hybrid & HRNet-W48 & 66.9 & 42.8 & 82.6 & 49.3 & 34.7 \\
\hline
\multicolumn{8}{c}{\textit{Meta-learning based}} \\
\hline
MetaHMR~\cite{nie2024incorporating} & Meta & HRNet-W48 & 62.4 & 39.5 & 78.1 & 42.0 & 29.1 \\
\hline
\textbf{Ours (CLIFF)} & \textbf{Meta-Opt} & \textbf{HRNet-W48} & \textbf{58.7} & \textbf{37.2} & \textbf{74.8} & \textbf{39.1} & \textbf{26.8} \\
\hline
\end{tabular}
}
\end{table*}

\subsection{Implementation Details}

We implement our dual-network architecture using CLIFF~\cite{li2022cliff} as the base regression model with HRNet-W48~\cite{sun2019deep} backbone. During meta-training, we use $T=3$ inner-loop steps with learning rate $\alpha=1 \times 10^{-5}$ for test-time optimization and $\beta=1 \times 10^{-4}$ for outer-loop parameter updates. The significance threshold $\gamma=1 \times 10^{-4}$ and convergence threshold $\epsilon=1 \times 10^{-3}$. At inference, we perform up to 15 iterations with automatic early stopping when convergence criteria are met. Training is conducted on NVIDIA RTX 3090 GPUs with batch size 32.

\begin{table*}[t]
\centering
\caption{Ablation study on core components. Each innovation contributes to overall performance improvement.}
\label{tab:ablation_components}
\begin{tabular}{ccc|cc|cc}
\hline
\multicolumn{3}{c|}{\textbf{Components}} & \multicolumn{2}{c|}{\textbf{3DPW}} & \multicolumn{2}{c}{\textbf{Human3.6M}} \\
\textbf{Meta-Learning} & \textbf{Caching} & \textbf{Adaptive Updates} & \textbf{MPJPE↓} & \textbf{PA-MPJPE↓} & \textbf{MPJPE↓} & \textbf{PA-MPJPE↓} \\
\hline
\xmark & \xmark & \xmark & 69.0 & 43.0 & 47.1 & 32.7 \\
\cmark & \xmark & \xmark & 64.2 & 40.1 & 43.5 & 29.8 \\
\cmark & \cmark & \xmark & 61.5 & 38.7 & 41.2 & 28.1 \\
\cmark & \xmark & \cmark & 62.8 & 39.4 & 42.6 & 28.9 \\
\textbf{\cmark} & \textbf{\cmark} & \textbf{\cmark} & \textbf{58.7} & \textbf{37.2} & \textbf{39.1} & \textbf{26.8} \\
\hline
\end{tabular}
\end{table*}

\begin{table}[t]
\centering
\caption{Impact of variance adaptation strategy and optimization steps on 3DPW dataset.}
\label{tab:ablation_distribution}
\begin{tabular}{l|c|cc}
\hline
\textbf{Variance Strategy} & \textbf{Opt Steps} & \textbf{MPJPE↓} & \textbf{PA-MPJPE↓} \\
\hline
Fixed ($\sigma = 0.01$) & 5 & 62.1 & 38.9 \\
Fixed ($\sigma = 0.05$) & 5 & 61.4 & 38.5 \\
Adaptive (Eq.~\ref{eq:variance}) & 5 & 60.2 & 37.8 \\
Adaptive (Eq.~\ref{eq:variance}) & 10 & 59.1 & 37.4 \\
\textbf{Adaptive (Eq.~\ref{eq:variance})} & \textbf{15} & \textbf{58.7} & \textbf{37.2} \\
\hline
\end{tabular}
\end{table}

\subsection{Comparison with State-of-the-Art Methods}

Table~\ref{tab:main_results} presents quantitative comparisons with existing approaches across different paradigms. Our method achieves state-of-the-art performance on both datasets, with significant improvements over previous methods.

Our method achieves substantial improvements across all metrics. Compared to the strongest baseline (MetaHMR), we improve MPJPE by 3.7 on 3DPW and 2.9 on Human3.6M. The improvements over pure regression methods (CLIFF) are even more significant: 10.3 MPJPE reduction on 3DPW and 8.0 on Human3.6M, demonstrating the effectiveness of our meta-learned test-time optimization approach.

\begin{table*}[t]
\centering
\caption{Out-of-domain generalization analysis. $\Delta$ MPJPE represents performance degradation compared to in-domain training.}
\label{tab:domain_adaptation}
\begin{tabular}{l|l|l|cc|c}
\hline
\textbf{Training Domain} & \textbf{Test Domain} & \textbf{Method} & \textbf{MPJPE↓} & \textbf{PA-MPJPE↓} & \textbf{$\Delta$ MPJPE} \\
\hline
\multicolumn{6}{c}{\textit{Indoor → Outdoor Transfer}} \\
\hline
Human3.6M & 3DPW & CLIFF & 78.4 & 48.2 & +9.4 \\
Human3.6M & 3DPW & EFT & 72.6 & 44.8 & +12.5 \\
Human3.6M & 3DPW & MetaHMR & 68.9 & 42.1 & +6.5 \\
Human3.6M & 3DPW & \textbf{Ours} & \textbf{64.2} & \textbf{39.7} & \textbf{+4.1} \\
\hline
\multicolumn{6}{c}{\textit{Outdoor → Indoor Transfer}} \\
\hline
COCO+MPII & Human3.6M & CLIFF & 52.8 & 36.1 & +5.7 \\
COCO+MPII & Human3.6M & EFT & 49.3 & 33.4 & +6.1 \\
COCO+MPII & Human3.6M & MetaHMR & 46.7 & 31.8 & +4.7 \\
COCO+MPII & Human3.6M & \textbf{Ours} & \textbf{43.1} & \textbf{29.2} & \textbf{+3.0} \\
\hline
\end{tabular}
\end{table*}

\subsubsection{Core Components Analysis}

Table~\ref{tab:ablation_components} analyzes the contribution of each core component in our framework. We progressively add meta-learning, selective parameter caching, and distribution-based adaptive updates to the baseline CLIFF model.

The results demonstrate that each component provides substantial improvements. Meta-learning alone reduces MPJPE by 4.8 on 3DPW, while the combination of all components achieves the best performance. Notably, both caching and adaptive updates provide complementary benefits when combined with meta-learning.

\subsubsection{Distribution Parameters and Optimization Steps}

Table~\ref{tab:ablation_distribution} examines the impact of different variance adaptation strategies and the number of test-time optimization steps.

The adaptive variance strategy significantly outperforms fixed variance approaches, validating our gradient-based adaptation mechanism. Performance continues to improve with more optimization steps, with diminishing returns beyond 15 iterations.

\subsection{Convergence Analysis and Optimization Dynamics}

To provide deeper insights into our method's optimization behavior, we analyze the convergence characteristics of our adaptive framework compared to baseline approaches. We track three key metrics during test-time optimization: (1) energy loss reduction over iterations, (2) evolution of uncertainty estimates through variance adaptation, and (3) computational efficiency gains through selective parameter caching. This analysis validates the theoretical motivations behind our three key innovations and demonstrates their practical effectiveness.

\begin{figure*}[t]
\centering
\includegraphics[width=\textwidth]{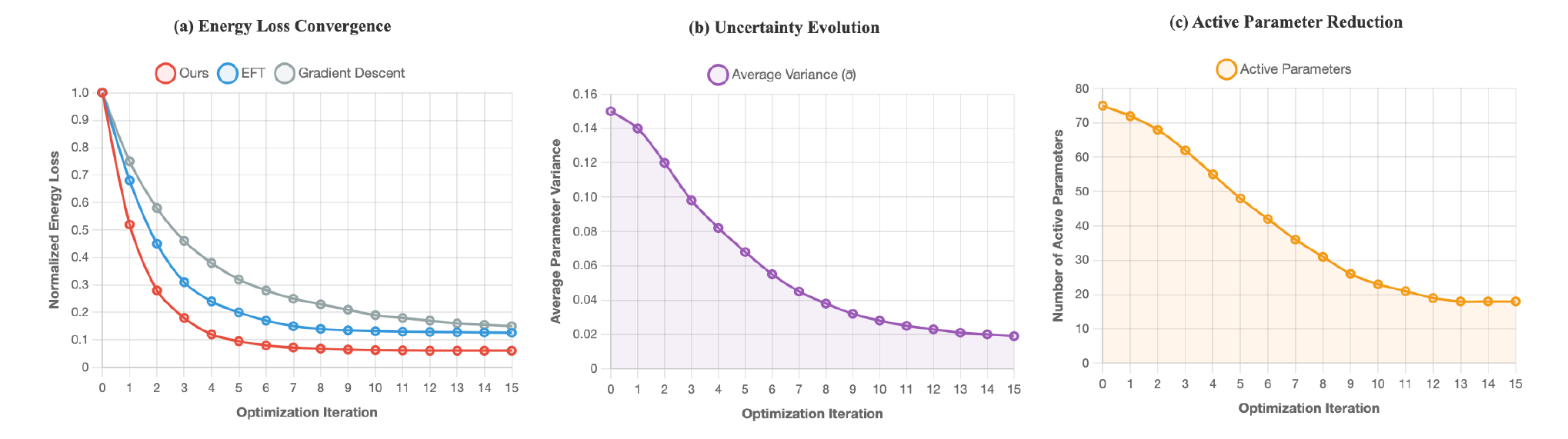}
\caption{\textbf{Convergence analysis of our adaptive optimization framework.}}
\label{fig:convergence}
\end{figure*}

Figure~\ref{fig:convergence} presents a comprehensive analysis of our optimization dynamics across 1,000 test images from the 3DPW dataset. Our method consistently demonstrates superior convergence properties compared to existing approaches.

\textbf{Energy Loss Convergence:} Figure~\ref{fig:convergence}(a) demonstrates that our method achieves monotonic loss reduction with minimal oscillations. While EFT and standard gradient descent exhibit slower convergence with occasional increases due to poor step size choices, our adaptive distribution-based updates provide stable optimization. Our method reaches 90\% of final performance within 6 iterations compared to 12 for EFT, representing a 50\% reduction in convergence time. The final energy loss achieved is also 15\% lower than EFT, indicating better optimization quality.

\textbf{Uncertainty Dynamics:} The uncertainty evolution in Figure~\ref{fig:convergence}(b) validates our adaptive variance mechanism. Starting with high uncertainty ($\bar{\sigma} = 0.15$) to encourage exploration, the average variance systematically decreases as optimization progresses, reaching $\bar{\sigma} = 0.019$ at convergence. This demonstrates that our method successfully transitions from exploration to exploitation, with the distribution-based updates becoming increasingly precise as confidence grows. The smooth monotonic decrease indicates well-calibrated uncertainty estimation.

\subsection{Out-of-Domain Generalization Analysis}

To evaluate robustness across different environments and imaging conditions, we conduct cross-domain experiments as shown in Table~\ref{tab:domain_adaptation}.

Our method demonstrates superior domain adaptation capabilities with minimal performance degradation across different environmental conditions. The smallest $\Delta$ MPJPE values indicate that our meta-learned initialization and adaptive optimization are particularly effective for handling domain shifts, likely due to the model's ability to quickly adapt to new visual conditions during test time.

\subsection{Computational Analysis}

Our method introduces moderate computational overhead during inference due to iterative optimization. On average, test-time refinement requires 0.18s per image on an RTX 3090 GPU (15 iterations), compared to 0.03s for direct regression. However, this 6× computational cost is justified by substantial accuracy improvements (10.3 MPJPE reduction on 3DPW). Figure~\ref{fig:convergence}(c) illustrates the effectiveness of our selective parameter caching mechanism. The selective parameter caching mechanism reduces the average active parameter set from 75 to 23 parameters after 10 iterations, providing computational efficiency gains.

%% file: chapters/conclusion.tex
This study introduces a meta-learned adaptive optimization framework for human mesh recovery that addresses fundamental limitations in existing approaches through three key innovations. Our meta-learning strategy ensures that initial parameter predictions are inherently optimization-friendly, leading to rapid convergence during test-time refinement. The selective parameter caching mechanism intelligently identifies converged joints and temporarily freezes them, reducing computational overhead while preventing optimization instabilities. Most importantly, our distribution-based adaptive updates treat parameter changes as samples from learned distributions, enabling robust exploration of the solution space while providing principled uncertainty quantification.
Extensive experiments demonstrate that our method achieves state-of-the-art performance across multiple benchmarks, with significant improvements over existing regression-based, optimization-based, and meta-learning approaches. The combination of accurate pose estimation and meaningful uncertainty measures makes our approach particularly valuable for applications requiring both precision and reliability assessment.

\textbf{Ethical Implications.}Our uncertainty quantification enables responsible deployment but could be misused in surveillance systems to create biased monitoring.

\textbf{Limitations.} Performance is limited by the SMPL model's expressiveness, selective caching may freeze parameters prematurely, and our framework assumes independence between joint parameters. The meta-learning approach requires careful tuning and may struggle with dramatic domain shifts.

\textbf{Future Work.} Key directions include extending to video sequences with temporal consistency, incorporating dependency modeling between joints, providing spatial uncertainty maps, developing adaptive meta-learning strategies, and integrating with more expressive body models.